# Automatic marker-free registration of tree point-cloud data based on rotating projection

Xiuxian Xu[a], Pei Wang[a,*], Xiaozheng Gan[a], Yaxin Li[a], Li Zhang[a], Qing Zhang[a], Mei Zhou[b], Yinghui Zhao[c], Xinwei Li[a]

[a] College of Science, Beijing Forestry University, Beijing 100083, China

[b] Key Laboratory of Quantitative Remote Sensing Information Technology, Academy of Opto-Electronics, Beijing 100094, China

[c] Aerospace Information Research Institute, Chinese Academy of Sciences, Beijing 100094, China

Corresponding author:

Pei Wang, wangpei@bjfu.edu.cn

**Abstract**

Point-cloud data acquired using a terrestrial laser scanner (TLS) play an important role in digital forestry research. Multiple scans are generally used to overcome occlusion effects and obtain complete tree structural information. However, it is time-consuming and difficult to place artificial reflectors in a forest with complex terrain for marker-based registration, a process that reduces registration automation and efficiency. In this study, we propose an automatic coarse-to-fine method for the registration of point-cloud data from multiple scans of a single tree. In coarse registration, point clouds produced by each scan are projected onto a spherical surface to generate a series of two-dimensional (2D) images, which are used to estimate the initial positions of multiple scans. Corresponding feature-point pairs are then extracted from these series of 2D images. In fine registration, point-cloud data slicing and fitting methods are used to extract corresponding central stem and branch centers for use as tie points to calculate fine transformation parameters. To evaluate the accuracy of registration results, we propose a model of error evaluation via calculating the distances between center points from corresponding branches in adjacent scans. For accurate evaluation, we conducted experiments on two simulated trees and a real-world tree. Average registration errors of the proposed method were 0.26m around on simulated tree point clouds, and 0.05m around on real-world tree point cloud.

# 1 Introduction

Three-dimensional (3D) geometric information describing trees is very important in many research fields for processes such as biomass estimation, forest inventory, forest management, and urban environment modeling (Dubayah and Drake, 2000; Popescu, Wynne and Nelson, 2003; Hopkinson *et al.*, 2004; Popescu, 2007; Wulder *et al.*, 2008). Some valid methods used to acquire tree structure information include traditional field measurement, photography, and laser scanning. In recent years, 3D laser scanners have been widely applied to acquire 3D tree information for different types of experiments.

Terrestrial laser scanner (TLS)-based methods have been developed to construct 3D models of trees for data extraction (Pfeifer *et al.*, 2004; Thies* *et al.*, 2004; Henning and Radtke, 2006; Dassot, Constant and Fournier, 2011; Raumonen *et al.*, 2013). Due to the geometric complexity of trees, TLS methods result in occlusion effects in each scan. This limitation leads to partial observation and incomplete structural information, which greatly increases the difficulty of fully reconstructing trees within a single scan. Reconstruction based on multiple scans is an efficient complementary method to mitigate occlusion effects and facilitate the full reconstruction of trees. Multiple-scan approaches produce point clouds from different scans that lie within different coordinate systems. Thus, multiple scans must be transformed to a common coordinate system via a registration procedure (Guiyun Zhou, Bin Wang and Ji Zhou, 2014).

Point-cloud registration methods can be categorized into two classes: marker-based and marker-free registrations. Marker-based registration relies on artificial markers that are manually placed at the scene and manual or automatic recognition of these markers in different scans to establish correspondences (Bienert and Maas, 2009; Hilker *et al.*, 2012). The markers are often reflective and can have various shapes (e.g., circular, cylindrical, or spherical). Based on corresponding point pairs extracted by identifying the same markers in adjacent scans, the relative transformation matrix between overlapping areas in multiple scans can be calculated by many commercial software packages to complete the registration procedure.

Marker-based registration is accurate and reliable but has many limitations. In complex environments, artificial markers can be difficult to place, and marker-based registration is often time-consuming in the field (Pfeifer *et al.*, 2004).

By contrast, marker-free registration attempts to automatically merge two or more scans directly without using artificial markers. Researchers using marker-free methods often focus on extracting natural geometric features (e.g., points, lines, and surfaces) from the scans (Böhm and Becker, 2007; Brenner, Dold and Ripperda, 2008). These features are utilized to extract tie points during registration. In forestry scenes, ground surface points, stem centers, and skeletons can be extracted to establish correspondence between multiple scans (Aschoff and Spiecker, 2004; Jason G. Henning and Radtke, 2008). The iterative closest point (ICP) algorithm and its variants are commonly used marker-based registration methods (Besl and McKay, 1992; Rusinkiewicz and Levoy, 2001). The ICP algorithm starts with two scans and an initial guess for relative rigid-body transformation; an iterative approach is then applied to refine the transformation by alternately establishing correspondence. Due to sensitivity to the initial position and the large computational cost of multiple iterations, ICP methods are often used in fine registration processes. Another important registration method is the four-point congruent set (4PCS), which extracts coplanar four-point sets from approximately congruent scans to complete global registration (Aiger, Mitra and Cohen-Or, 2008). Without the requirement of assumptions about initial alignment, 4PCS can establish reliable corresponding sets within a limited number of trials and is robust against noise and low-overlap scans.

In forest scenes, the complex geometric distribution of branches and large number of leaf points pose a challenge to maker-free registration of tree point-cloud data (Bailey and Ochoa, 2018). Trees typically have a symmetric geometric structure, making automatic registration more difficult. Because we cannot guarantee the simultaneous acquisition of multiple scans, natural elements (e.g., wind, sun, and animals) will introduce inconsistencies in overlapping parts among multiple scans. In most situations, leaf points will interfere with accurate registration; a few methods have been proposed to solve this problem. (Jason G Henning and Radtke, 2008) used tie points estimated from ground surfaces and stem centers in range images to register forestry scenes. The process of extracting tree stems in the method is not free of manual steps. (Bucksch and Khoshelham, 2013) applied localized registration using a skeletonization method to detect correspondences between branch segments in multiple scans. However, this approach relied on roughly registered tree point-cloud data prior to fine registration. (Guiyun Zhou, Bin Wang and Ji Zhou, 2014) applied a skeleton extraction method to a rough automatic registration procedure—based on the extracted skeleton, the initial translation vector and rotation angle were estimated using root point positions, distances between branch segments, and a mapping cost function between skeletons. By minimizing the mapping cost function, the transformation parameter was further refined in fine registration.

Recently, (Zhang *et al.*, 2016) proposed a coarse-to-fine strategy to address the difficulty of forestry scene registration. In coarse registration, a backsighting orientation procedure is used to calculate transformation parameters instead of placing artificial reflectors. Based on the initial values, stem-center locations are extracted as tie points to refine the rigid-body transformation for fine registration. The coarse-to-fine strategy improves the robustness and accuracy of forest scene registration, but also has several limitations. First, coarse registration requires manual placement of backsighting reflectors, which can be difficult to apply in complex environments. Second, due to the features of stem-fitting methods, fine registration cannot guarantee high registration accuracy in the vertical direction, especially for bent trunks whose cross sections cannot be treated as circles; the stem-fitting approach fails in such situations.

The registration of single-tree point-cloud data without reflectors remains a challenge and can be more difficult than that of a forest scene. Unlike multiple tree registration in a forest scene, where the spatial relationship between trees can be useful information, geometrical structure is the only information that can be used in marker-free registration of single-tree point-cloud data.

The objective of this study was to develop a fully automatic marker-free registration algorithm with high registration accuracy. A coarse-to-fine registration strategy was adopted to align point clouds with bad initial positions without reference points in a stepwise manner. In our coarse registration, each 3D point cloud was projected onto a sphere to generate a series of 2D projection images for the extraction of feature-point pairs, whose spatial information can be used to estimate the transformation matrix for the coarse registration of multiple scans. Sliced point-cloud data were then used to estimate the centers of trunks and branches using fitting methods. The estimated centers were used as tie points to perform fine registration.

This paper is organized as follows. Section 2 explains the workflow of the proposed coarse-to-fine registration method in detail. Section 3 presents the experimental results based on the data of two simulated trees using the proposed method and compares these results with those obtained using the ICP algorithm. In addition, the experiment results on a real-world tree point cloud are presented. Section 4 discusses the method and suggests improvements. Section 5 presents the conclusions of the study and directions for future work.

**2 Methodology**

In our proposed method, the registration procedure comprises two parts: coarse and fine registration. The objective of the coarse-to-fine strategy is to transform coordinates from target points to reference points in a stepwise manner. In coarse registration, a rough transformation matrix is calculated to transform the target points into a position that is close to the reference points. Based on the close relative positions of the two point sets, more information can be used to achieve an accurate transformation towards the reference points in fine registration. In both steps, rigid-body transformation is determined by translation and rotation parameters. The registration procedure is described by following equations:

$$pt_{coarse} = R_{coarse} \cdot pt_{tar} + T_{coarse}$$

$$pt_{ref} = R_{fine} \cdot pt_{coarse} + T_{fine} \quad (1)$$

Where $pt_{tar}$ and $pt_{ref}$ are points in the target and reference scans, respectively; $R$ is the rotation matrix; and $T$ is the translation vector.

In coarse registration, a dimension-reduction method simplifies the point-matching problem by projecting from the 3D point cloud to 2D images. The matched points are estimated using feature-point-matching algorithms on images generated by projection. These matched points are then used to estimate the rough transformation.

In fine registration, point-cloud slicing and fitting methods are used to extract corresponding central stem centers and branch centers, which function as tie points to calculate the fine transformation parameters. The workflow is illustrated in Fig. 1.

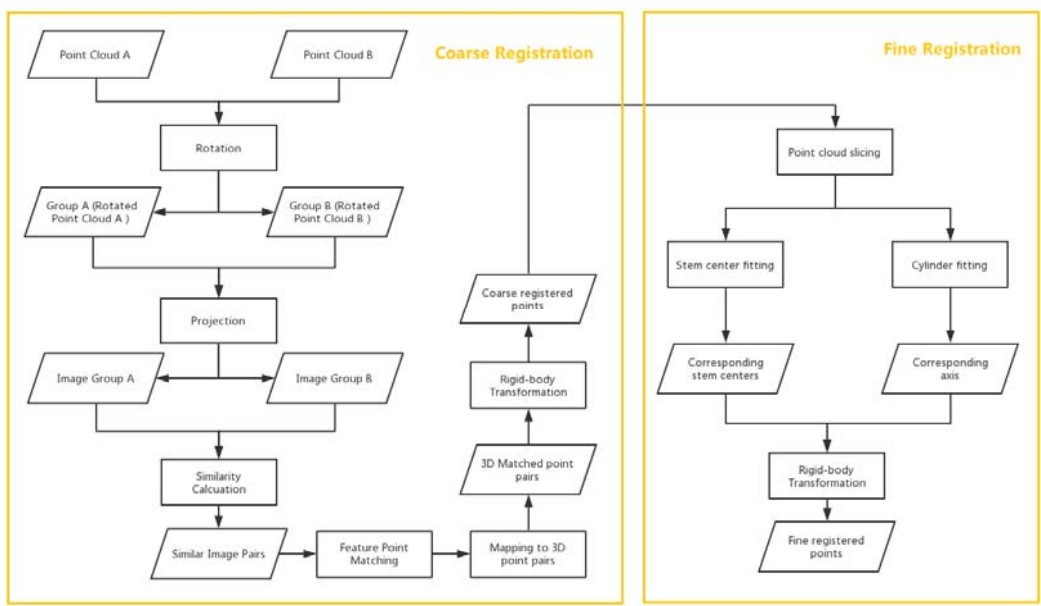

Fig. 1. Workflow of the coarse-to-fine registration procedure.

2.1 Coarse registration

2.1.1 Point-cloud projection

We established a point-cloud projection model to convert 3D point clouds to 2D images. The tree point-cloud is projected onto a sphere centered at the origin of the coordinate system, where the scanner is located. The projection was then used to generate an image on the spherical surface. The model is shown in Fig. 2.

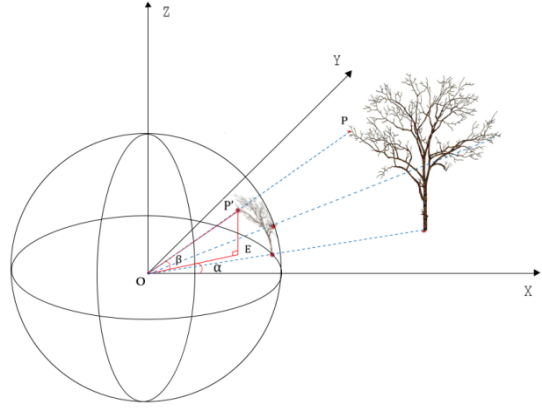

Fig. 2. The point-cloud projection model. P' is the projection of point P on the spherical surface. E is the foot of the perpendicular line from P' to the X-Y plane. α is the angle between OE and the x-axis. β is the angle between OP' and the X-Y plane.

Each point projected onto the sphere corresponds to a pair of angles, α and β (Fig. 2), which are used to determine the pixel coordinates of each point and generate corresponding images.

In the model, each set of pixel coordinates corresponds to a pair of intervals [$α_l$, $α_h$] and [$β_l$, $β_h$] for angles α and β respectively. The pixel coordinates of a point is determined by which pair of intervals its corresponding angles α and β lie in. Each point is allocated a pixel coordinate in this manner. The minimum steps of angle α and β are equal to the horizontal and vertical angular step widths of the TLS instrument, denoted φ and ϕ, respectively. In this study, each pixel in the image covered a region of 2φ and 2ϕ in the horizontal and vertical directions, respectively, which reduced the impact of discontinuous scanning points. The projected image has a size of m × n, where the values of m and n are calculated using the following equations:

$$m = \left| (α_{max} - α_{min}) / 2φ + 2r_1 \right|$$

$$n = \left| (β_{max} - β_{min}) / 2ϕ + 2r_2 \right| \quad (2)$$

where $α_{min}$, $α_{max}$, $β_{min}$, and $β_{max}$ are the minimum and maximum values of α and β of all points, respectively; and $r_1$ and $r_2$ are pixels forming a border around the image, which ensures that the size of the image satisfies our demands.

For any point P with corresponding angles $α_p$ and $β_p$, the pixel coordinates (x, y) can be calculated as follows:

$$x = (\alpha_p - \alpha_{\min}) / 2\varphi + r_1$$

$$y = (\beta_p - \beta_{\min}) / 2\phi + r_2 \qquad (3)$$

As a result, each scanning point corresponds to a certain set of pixel coordinates, and each pixel may have several corresponding points. The image generated via projection is a binary image. Pixels with and without corresponding scanning points are set to values of 0 and 255, respectively.

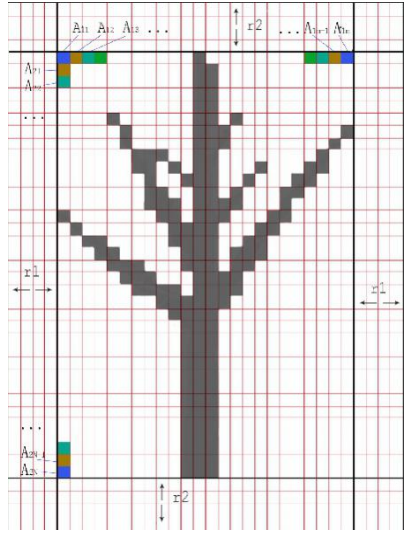

Fig. 3. Sample image generated via projection.

As shown in Fig. 3, $A_{ij}$ ($i = 1, 2, \ldots, m - 2r$; $j = 1, 2, \ldots, n - 2r_2$) is a pixel that corresponds to a pair of angle intervals. For example:

$A_{11} : ([\alpha_{\min}, \alpha_{\min} + \Delta\alpha), [\beta_{\min}, \beta_{\min} + \Delta\beta)), A_{12} : ([\alpha_{\min} + \Delta\alpha, \alpha_{\min} + 2\Delta\alpha), [\beta_{\min}, \beta_{\min} + \Delta\beta)), \ldots, A_{1,m-2r_1} : ([\alpha_{\max} - \Delta\alpha, \alpha_{\max}), [\beta_{\min}, \beta_{\min} + \Delta\beta))$

$A_{21} : ([\alpha_{\min}, \alpha_{\min} + \Delta\alpha), [\beta_{\min} + \Delta\beta, \beta_{\min} + 2\Delta\beta)), A_{31} : ([\alpha_{\min}, \alpha_{\min} + \Delta\alpha), [\beta_{\min} + 2\Delta\beta, \beta_{\min} + 3\Delta\beta)), \ldots, A_{n-2r_2,1} : ([\alpha_{\min}, \alpha_{\min} + \Delta\alpha), [\beta_{\max} - \Delta\beta, \beta_{\max}))$

2.1.2 Generation of image sequences

According to the projection method, projected images of the same object may differ due to differences among viewpoints. In our model, the projection viewpoint is determined by the position of the scanner. Due to occlusion effects, some valuable tree structural information will be lost in the process of dimension reduction. Thus, two scans with different viewpoints may be similar in 3D space, but their projected images may differ greatly, which can be an obstacle in identifying corresponding points between scans.

To solve this problem, we continuously rotated the tree point-cloud in 3D space prior to projection, which is equivalent to continually changing the viewpoint. In the rotation step, the mean values $\bar{x}$, $\bar{y}$ of the X and Y coordinates of all points in the scan were first calculated. We then continuously rotated the points by a certain degree around the axis, which is perpendicular to the *xy* plane and passes through the point ($\bar{x}$, $\bar{y}$, 0). As a result, a sequence of images was generated for each scan (Fig. 4).

The number of rotations required is often determined by the number of scans *n* and the rotation degree θ. In our experiments, the rotation degree θ was typically 10° and the rotation number was $\dfrac{720}{n \times \theta}$ (rotation obtained from $-\dfrac{360}{n}$ to $\dfrac{360}{n}$).

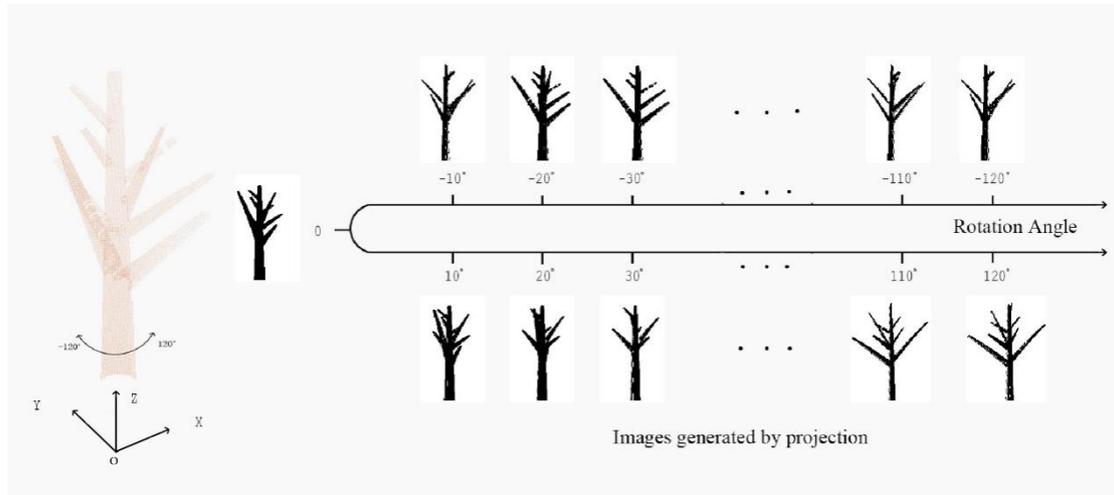

Fig. 4. Generation of image sequences.

2.1.3 Feature-point matching

Due to the lack of detailed texture information available in binary images, we used the ORB (oriented FAST and rotated BRIEF) algorithm to detect and match feature points (Rublee *et al.*, 2011). This method is faster and more suitable for less complicated images than methods such as the SURF (speeded up robust feature) or SIFT (scale invariant feature transform) algorithms (Lowe, 2004; Bay, Tuytelaars and Van Gool, 2006). The combination of the oriented FAST key point detector and rotated BRIEF descriptor makes the ORB algorithm scale- and rotation-invariant.

The key points in three pairs of similar images from adjacent scans were extracted and described using ORB. In each image pair, we selected the five pairs of matching points with the highest scores.

2.1.4 Transformation calculations

Once the matching points in images were determined, we were able to map the points to their corresponding 3D points in the tree point-cloud. First, we determined the intervals, [$\alpha_l$, $\alpha_h$] and [$\beta_l$, $\beta_h$], of each matching point based on its pixel coordinates. All scanning points with a corresponding pair of angles ($\alpha$, $\beta$) within these intervals were then extracted. The central point O among the extracted points was calculated and used as the tie point in 3D space. After obtaining more than four pairs of tie points in adjacent scans, a rough rigid-body transformation matrix between scans was calculated using singular-value decomposition (SVD) (Challis, 1995).

2.2 Fine registration

Coarse registration roughly aligns the postures of adjacent scans and provides a better initial position for subsequent fine registration. However, dimension reduction during coarse registration decreases the accuracy of registration. Obvious dislocation and separation remain between adjacent scans after coarse registration. Therefore, it is necessary to improve the transformation matrix in fine registration. The fine registration process includes three parts: point-cloud slicing, point separation, and circle and cylinder fitting.

2.2.1 Point-cloud slicing

Qualified tie points are the basis for the calculation of accurate transformation parameters in fine registration. For a single scan of a single tree, the point cloud is incomplete, and it is difficult to find tie points directly based on tree structure. In our method, we sliced points from the stem and branch parts of the tree and extracted the center points of stems and branches for use as tie points by applying circle- and cylinder-fitting methods to the sliced points. We sliced the points at quartiles of tree height and obtained three layers of points. Each layer was sliced to a thickness of 10 cm (Fig. 5).

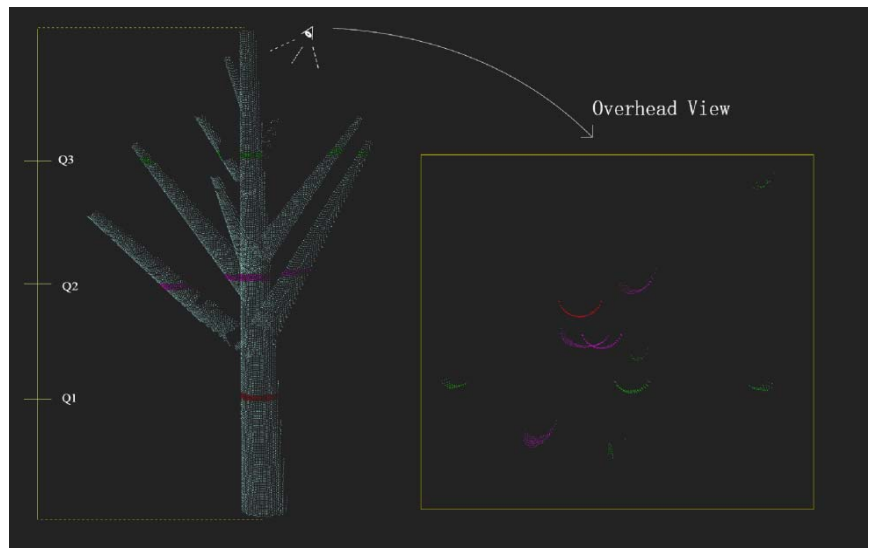

Fig. 5. Sliced points from a tree point-cloud. Q3, Q2, and Q1 are quartiles of tree height.

Different slicing heights result in differences in the number, distribution, and relative position of the sliced points in the layer. Points in lower layers are generally all extracted from the trunk. However, most points in high layers are extracted from branches. The trunk is usually perpendicular to the ground and its horizontal cross section is roughly treated as a circle. Thus, the center of the trunk can be estimated via circle fitting. Branches are often at an oblique angle to the trunk and their horizontal sections are similar to ellipsoids. Given that the geometric structure of branches in 3D space is similar to that of cylinders, we used a cylinder-fitting method to estimate the center points of the branches.

2.2.2 Point separation

Every sliced-point layer contains several arcs of points corresponding to branches and the trunk. Before applying fitting methods, we should first separate these arcs of points. Because the angular step of the TLS is fixed, whether two points are consecutive can be judged from the distance between their corresponding angles α and β (Fig. 1). Based on the horizontal and vertical angular step widths—$\varphi$ and $\phi$—of the TLS instrument, we separated the points by judging their connectivity (Bu and

Wang, 2016). The points in an arc are consecutive, and the distance between the angles—Δα and Δβ— corresponding to each pair of adjacent points should equal φ and ϕ, respectively, under ideal conditions (Fig. 6c). Because discontinuity can be caused by scanning errors or unusual tree structures, we determined the consecutive nature of two points by comparing Δα and Δβ to 3φ and 3ϕ, respectively. By examining distances between points, the method identified all connected areas and separated all sliced-point arcs (Fig. 6).

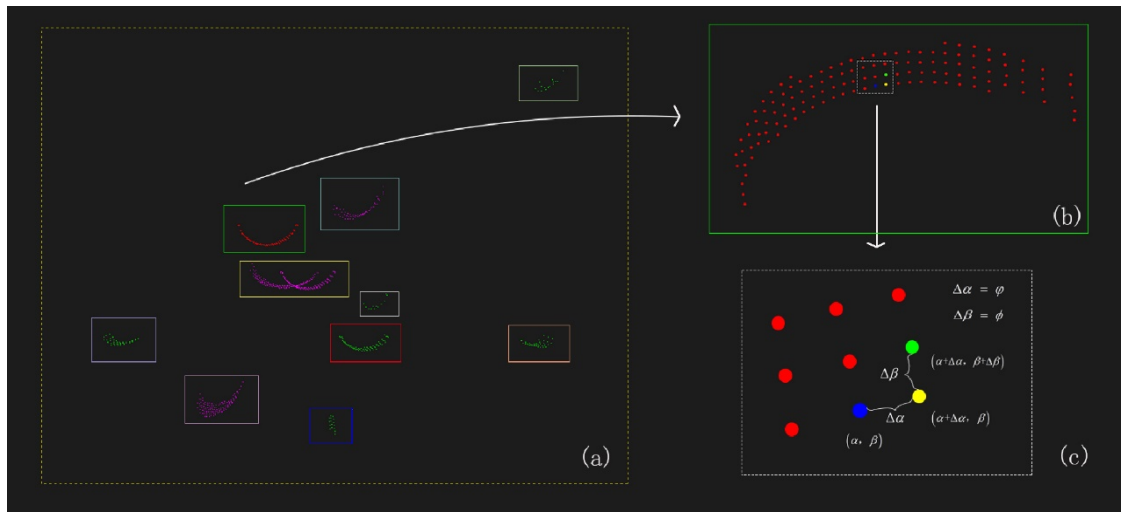

Fig. 6. Separation results. (a) Overview. (b) Close-up view of the connected part (within green rectangular box). (c) Close-up view of nine adjacent points in the connected part. Blue, yellow, and green points correspond to the angle pairs (α, β), (α + Δα, β), and (α + Δα, β + Δβ), respectively.

2.2.3 Circle and cylinder fitting

Among the three sliced layers, the lowest layer usually contains only one arc, corresponding to the trunk. For the trunk section, the circle-center position ($X_0$, $Y_0$) was extracted as the center of the trunk by applying the Taubin method (Taubin, 1991).

For sections of branches in higher layers, we determined center points by cylindrical fitting based on the least squares method (Shakarji, 1998). In the fitted cylinder, we obtained the direction vector of the central axis $(a, b, c)$, a starting point on the axis $(x_0, y_0, z_0)$, and radius R.

The axis of the branches could describe the tree to some extent (Eysn *et al.*, 2013). Thus, the starting point and a point one distance unit away from it in the positive direction of the axis were used as tie points. The positive direction of the axis was defined as the direction in which the Z coordinate of the point increases. For trunk points, the center of the fitted circle was regarded as a tie point.

As a result, we obtained a group of corresponding tie points with positions at different positions in the tree. Based on these tie points, transformation parameters were calculated for fine adjustment in fine registration.

## 3 Experimental results

Two simulated trees were used to verify our methods. Each tree was scanned three times (Fig. 7). Compared with real trees, the simulated trees had simpler geometric structure and less noise, which is useful for analyzing the advantages and disadvantages of our method.

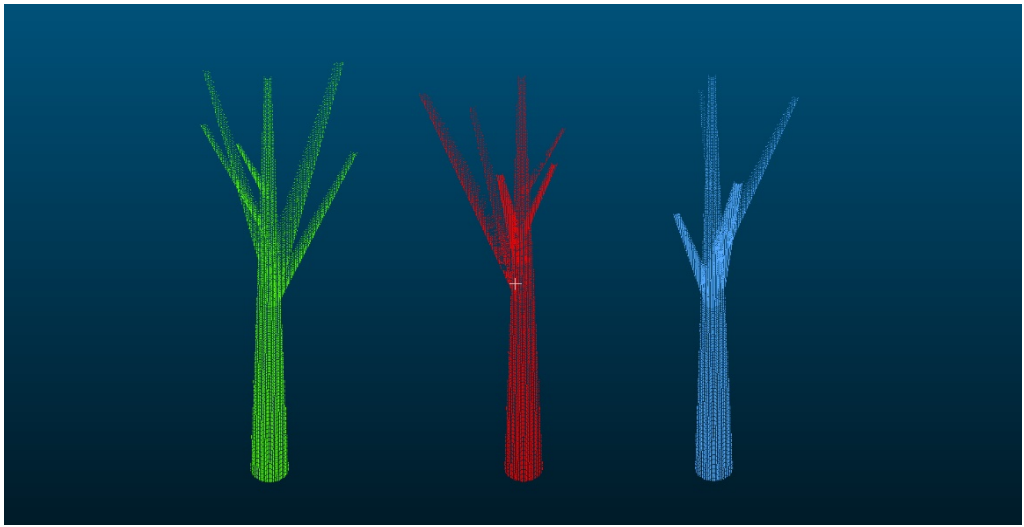

(a)

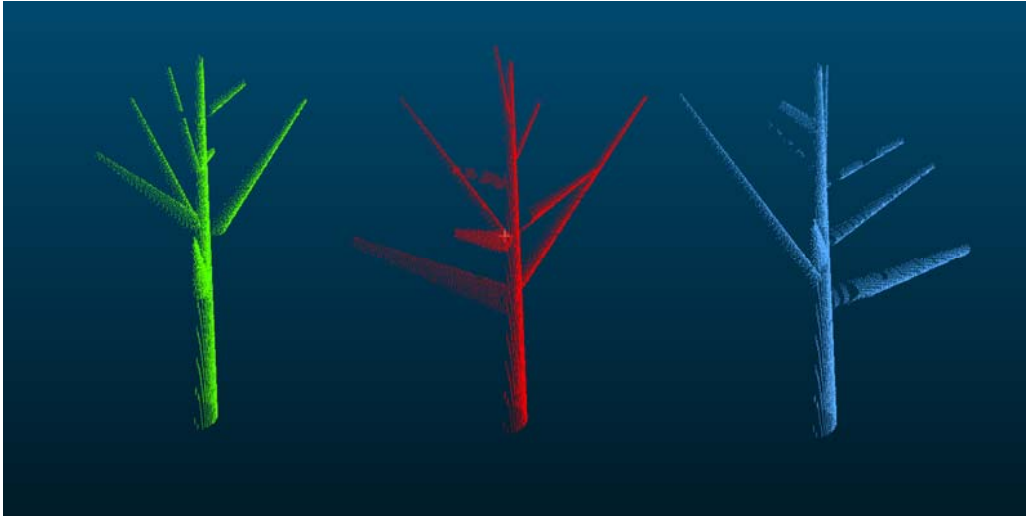

(b)

Fig. 7. Simulated point clouds for trees (a) A and (b) B. Green, red, and blue points indicate scans 1, 2, and 3, respectively.

3.1 Coarse registration

Each simulated tree was composed of three scans (Fig. 3). In the registration procedure, the local coordinate system of the first scan of each simulated tree was used as a reference coordinate system; the registration order was scan 2 to scan 1 and scan 3 to scan 1.

Similar image pairs between image sequences from adjacent scans were matched; the results of feature-point matching for two simulated trees are shown in Fig. 8. In each group of adjacent scans, three similar image pairs were selected for application of the ORB algorithm; 15 matching points were obtained.

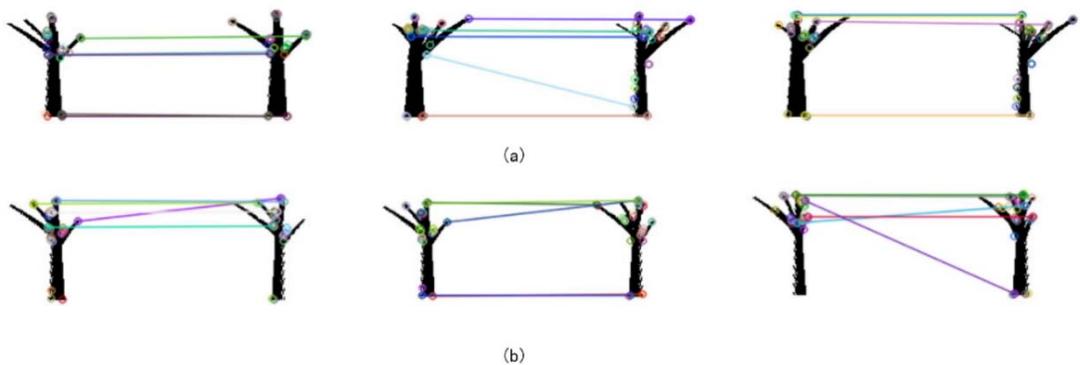

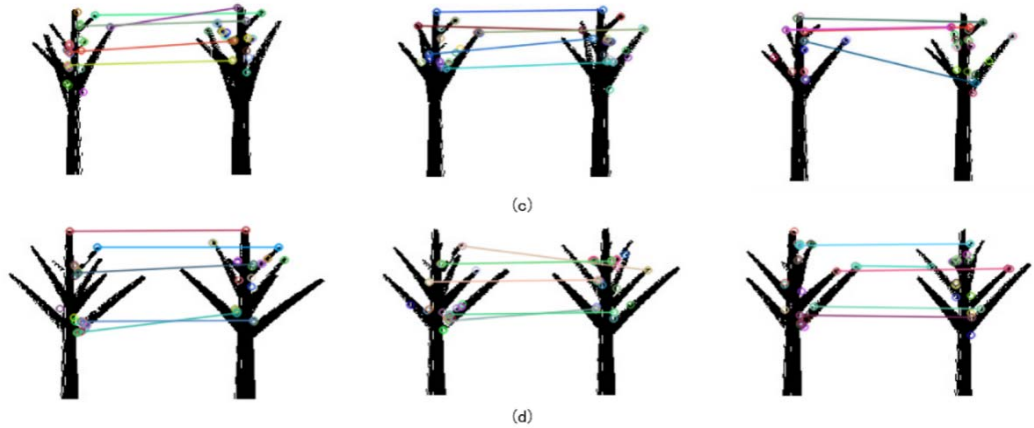

Fig. 8. Feature-point matching results. (a) Results of scan 2 to scan 1 for simulated tree A. (b) results of scan 3 to scan 1 for simulated tree A. (c) Results of scan 2 to scan 1 for simulated tree B. (d) Results of scan 3 to scan 1 for simulated tree B.

The coarse registration results are shown in Fig. 9. Corresponding trunks and branches between adjacent scans of both simulated trees were either crisscrossed or separate after coarse registration; i.e., not accurately aligned. Although registration errors cannot be ignored, coarse registration correctly matched corresponding trunks and branches between adjacent scans and transformed the target scan to a good initial position for fine registration. Coarse registration is also completely automatic and marker-free, which expands the potential fields for its application.

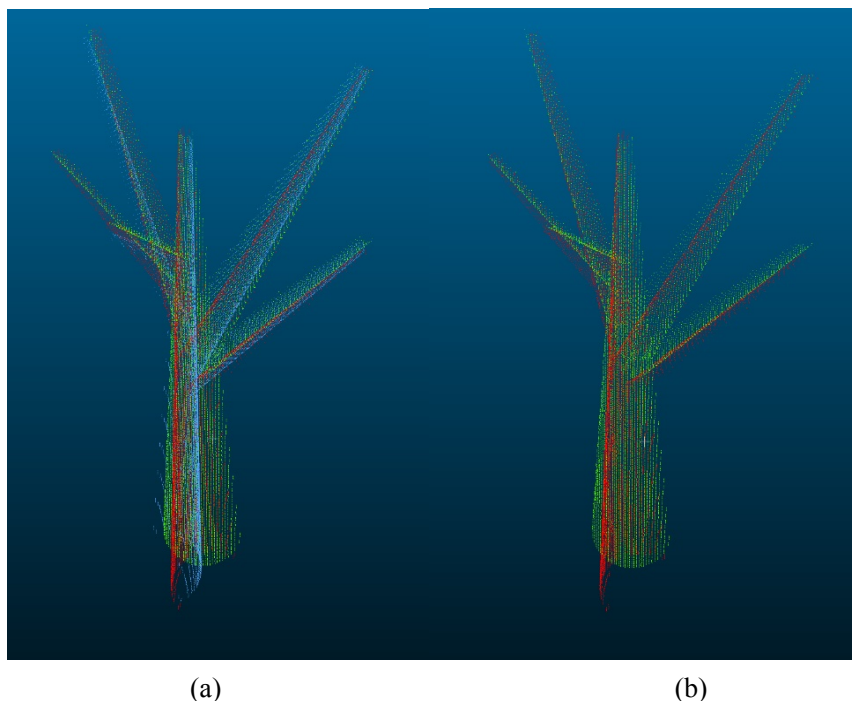

(a)                    (b)

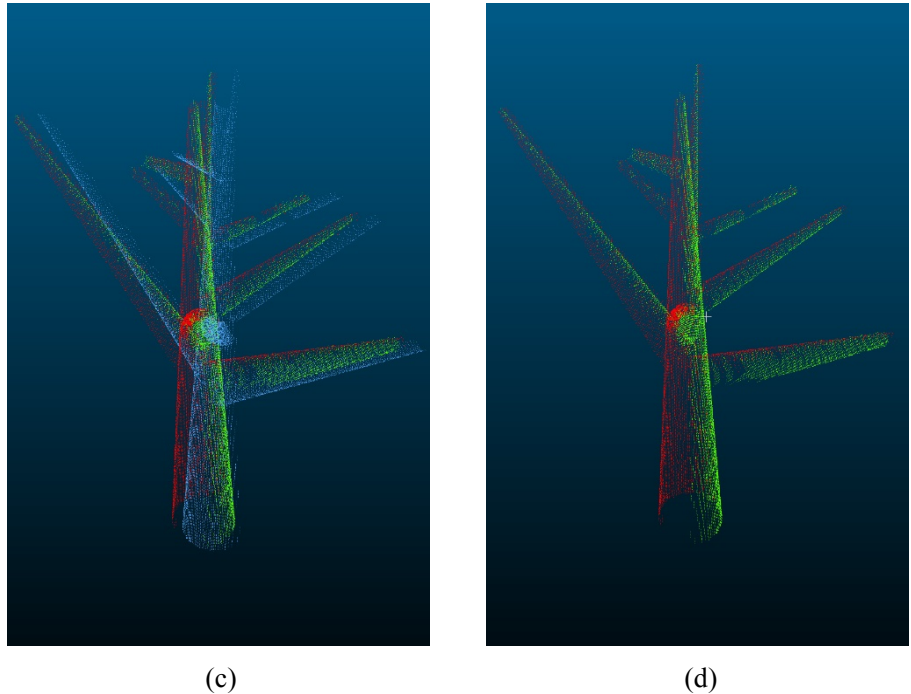

(c)                                  (d)

Fig. 9. Coarse registration results. (a) Registration result between scan 2 and scan 1 for simulated tree A. (b) Registration results between scan 3 and scan 1 added to results from (a). (c) Registration results between scan 2 and scan 1 for simulated tree B. (d) Registration results between scan 3 and scan 1 added to results from (c). Green, red, and blue points indicate scans 1, 2, and 3, respectively.

3.2 Fine registration

Based on target-scan initial positions, the center points of corresponding trunks and branches between adjacent scans were extracted as tie points to achieve better alignment in fine registration. Fine registration results for the two simulated trees are shown in Fig. 10. The results indicate that contours of the trunk and branches were complete and that a complete simulated tree could be composed from three scans. Alignment between adjacent scans was more accurate after fine registration.

Because multiple layers at different heights are sliced to facilitate the extraction of corresponding tie points, fine registration not only enhances the accuracy of coarse registration, but also achieves better alignment of branch and trunk parts between adjacent scans than fine registration via stem-center fitting methods.

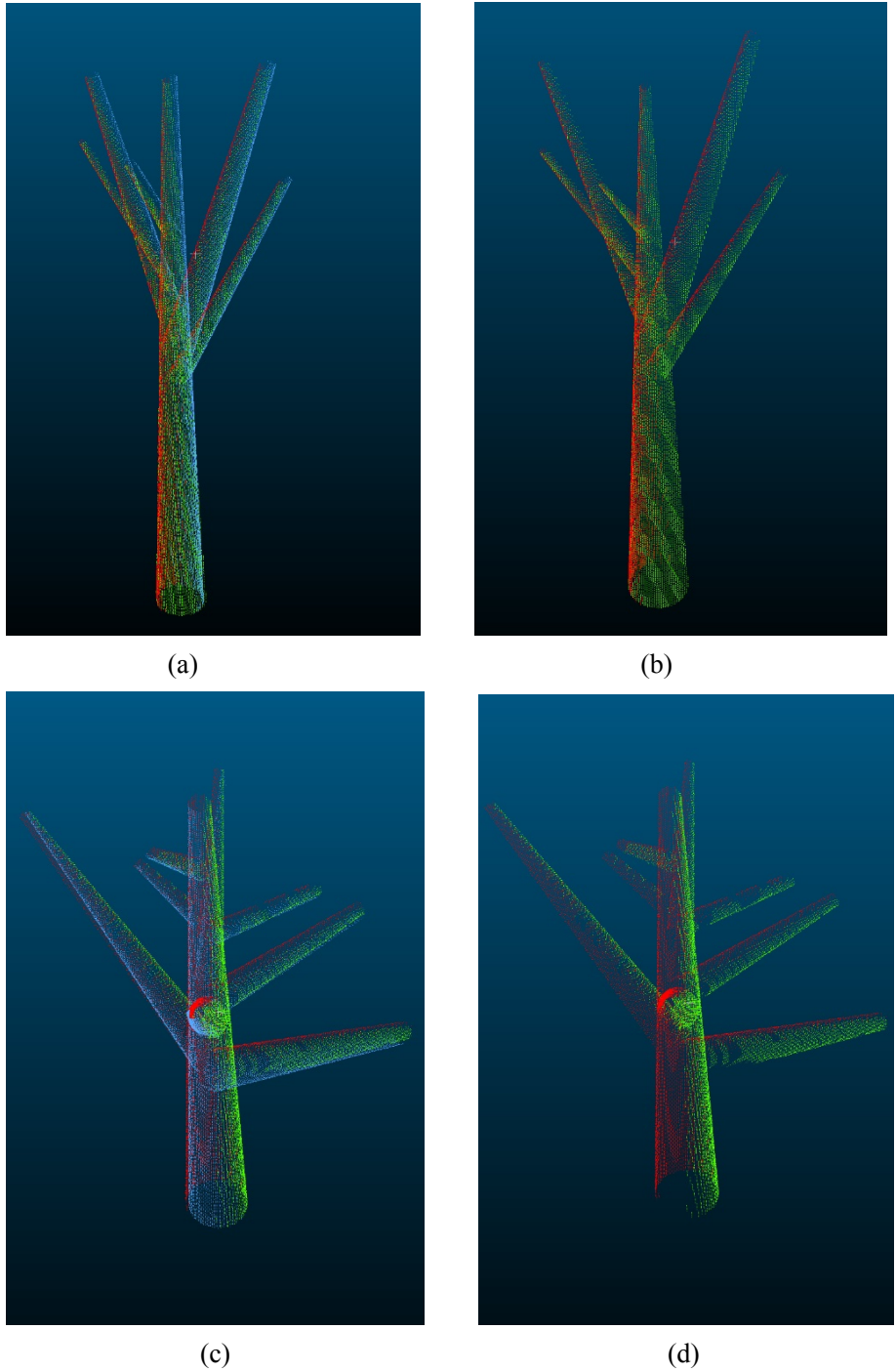

Fig. 10. Fine registration results. (a) Registration results between scan 2 and scan 1 for simulated tree A. (b) Registration results between scan 3 and scan 1 added to results from (a). (c) Registration results between scan 2 and scan 1 for simulated tree B. (d) Registration results between scan 3 and scan 1 added to results from (c). Green, red, and blue points indicate scans 1, 2, and 3, respectively.

3.3 Evaluating registration results

To evaluate our registration results, we developed an evaluation model to quantitatively estimate registration accuracy. Point clouds of simulated Trees A and B contained many branch parts (Fig. 11). Accurate registration would align corresponding branches between adjacent scans; the cross-section of a branch in a well-registered tree should be an ellipse or a circle. However, poor registration often results in branches that appear to be aligned correctly, but have cross-sections composed of several separate arcs. Thus, the alignment accuracy of corresponding branches between adjacent scans can be used to evaluate registration accuracy.

In our evaluation model, we extracted corresponding branches between adjacent scans to calculate registration error. For each pair of corresponding branches, we sliced three pairs of layers from the bottom, middle, and top of the corresponding branches. Points in these layers were used for cylinder fitting to estimate their center axes. By extracting points with the same z-values on the corresponding axes of each layer pair, we obtained three pairs of corresponding center points (Fig. 12). Calculating the distances between these pairs of center points can facilitate the estimation of registration error between adjacent scans, as follows:

$$\overline{d} = \frac{d_1 + d_2 + d_3 \cdots + d_n}{n} \qquad (8)$$

Where $d_1$, $d_2$, …, $d_n$ are the distances of corresponding center points, n is the number of pairs of corresponding points, and average distance $\overline{d}$ is treated as registration error.

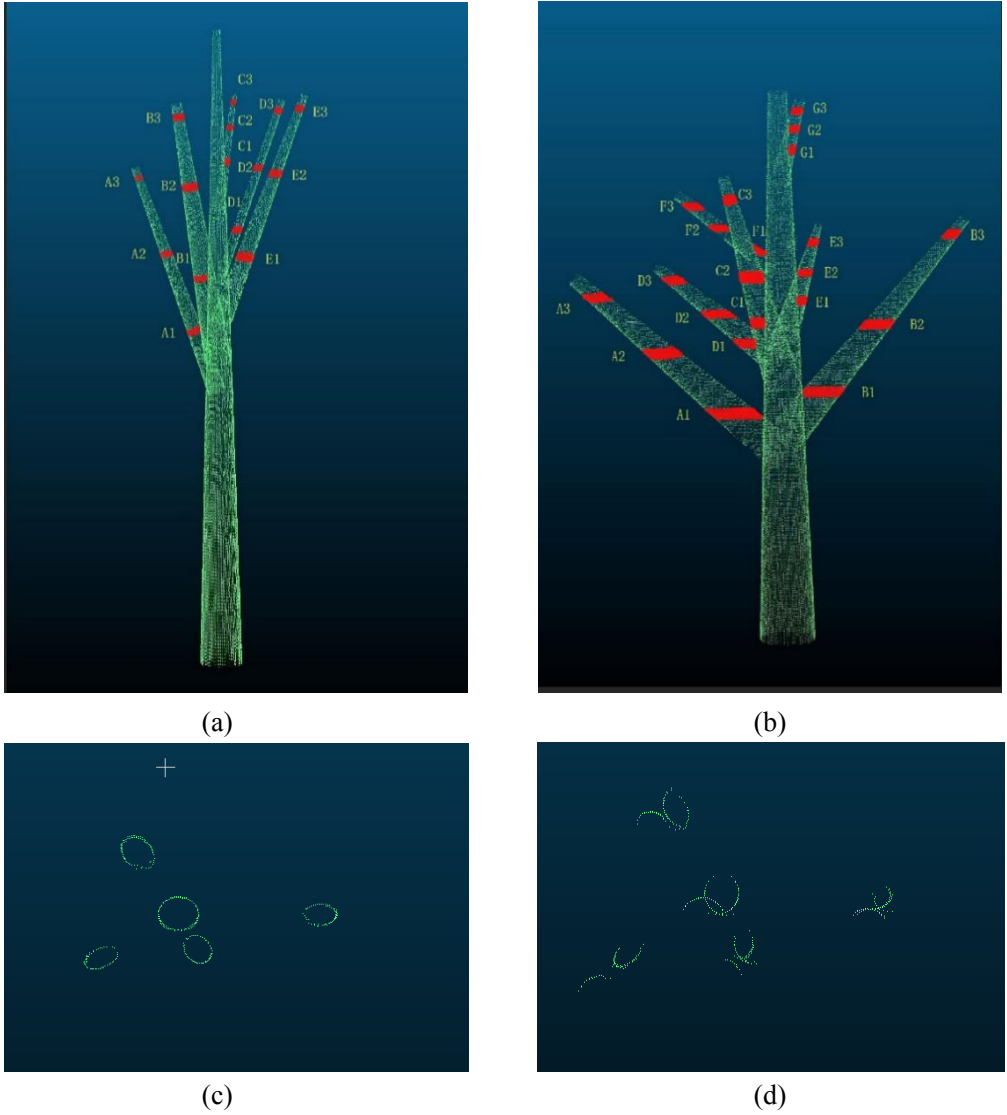

Fig. 11. Complete simulated tree point-clouds for the experiment. (a) Simulated tree A. (b) Simulated tree B. Red, sliced branch parts. (c) Cross-sections of branches at a given height in a well-registered tree point-cloud. (d) Cross-sections of branches at a given height in a poorly registered tree point-cloud.

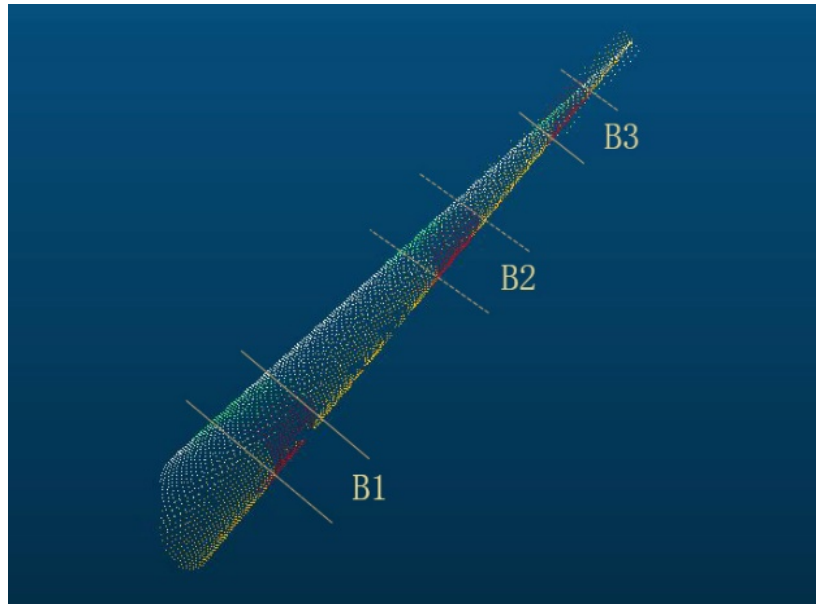

(a)

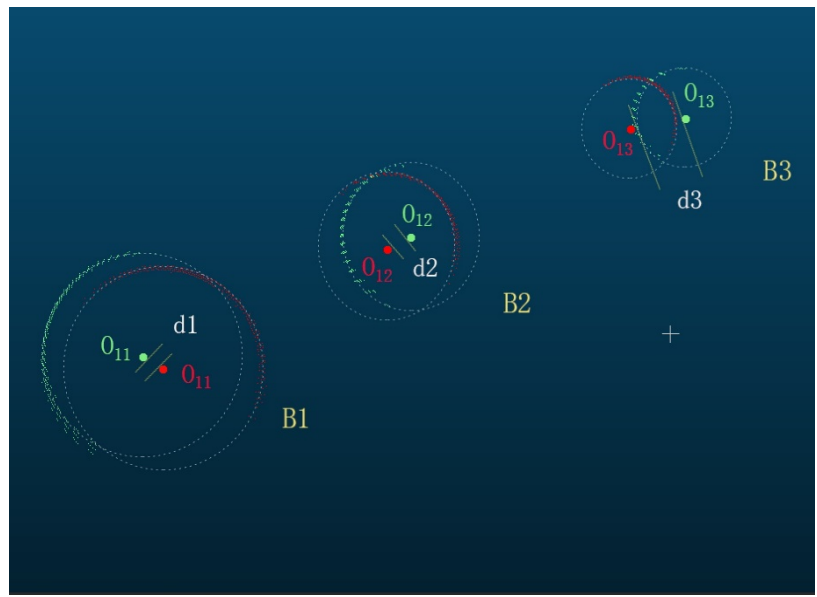

(b)

Fig. 12. Slicing results for one branch. (a) Three sliced layers of branch B in simulated tree B. White and yellow points indicate scans 1 and 2, respectively; green and red points indicate sliced points in scans 1 and 2, respectively. (b) Corresponding center-point extraction results for the three sliced layers of branch B. $d_1$, $d_2$, and $d_3$ indicate distances between the pairs of corresponding points.

To evaluate the registration accuracy of the two simulated trees, all corresponding branches between adjacent scans were extracted to calculate the corresponding center points and their distances. Branch section numbers for simulated trees A and B are shown in Fig. 11a and b. For comparison, the ICP algorithm was used in fine registration. As the ICP algorithm has high computational demand for the initial positions of adjacent scans, the algorithm was used only for fine registration in our experiment to ensure that the comparison was meaningful. The accuracies of the coarse and fine registration for each simulated tree are shown in Figs. 13 and 14. Table 2 shows the calculated mean registration errors.

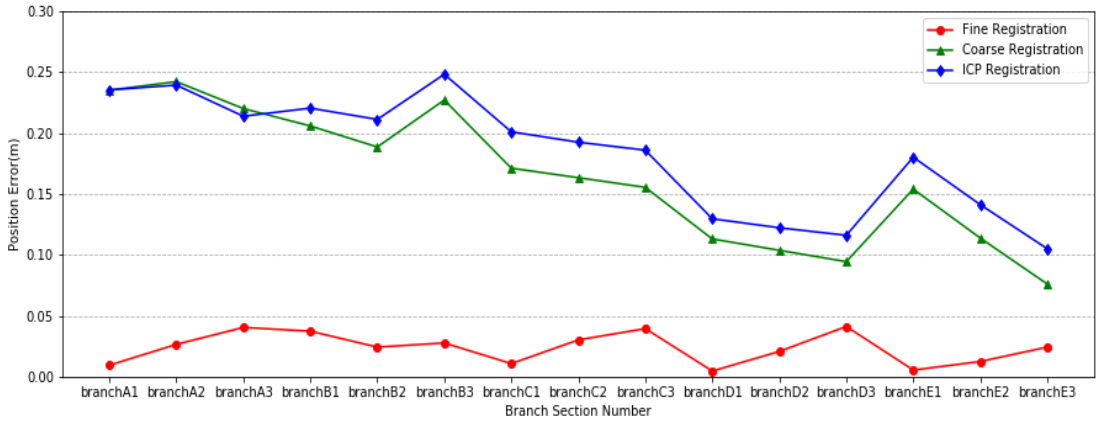

(a)

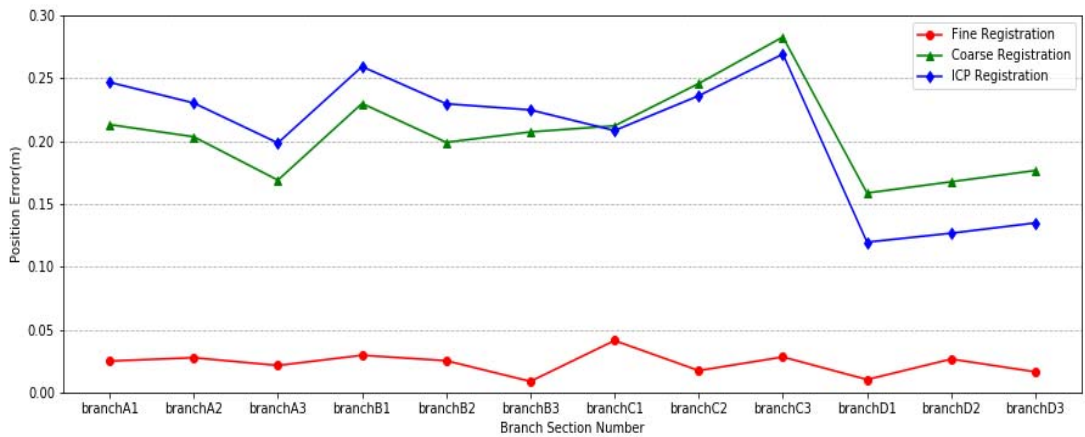

(b)

Fig. 13. Registration accuracy for simulated tree A. (a) Registration accuracy for scan 2 to scan 1. (b) Registration accuracy for scan 3 to scan 1.

Table 2  The errors of registration of simulated tree A and simulated tree B.

| Registration errors(m) | Coarse registration | Fine registration | ICP registration |
|---|---|---|---|
| Simulated Tree A  2-1 | 0.164 | 0.024 | 0.183 |
| Simulated Tree A  3-1 | 0.205 | 0.023 | 0.207 |
| Simulated Tree B  2-1 | 0.122 | 0.028 | 0.062 |
| Simulated Tree B  3-1 | 0.324 | 0.030 | 0.296 |

Note：2-1 means the registration order "Scan 2 to Scan 1", 3-1 means the registration order "Scan 2 to Scan 1".

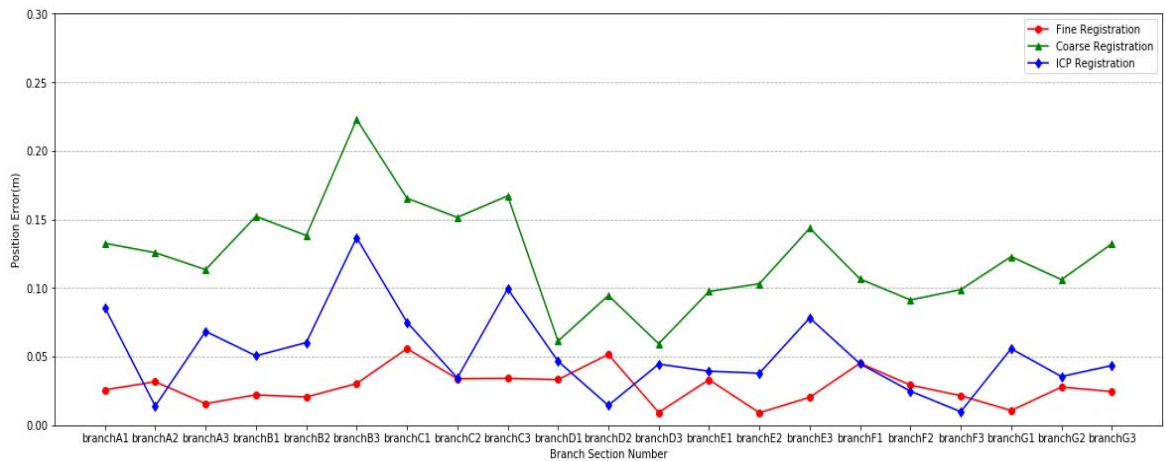

(a)

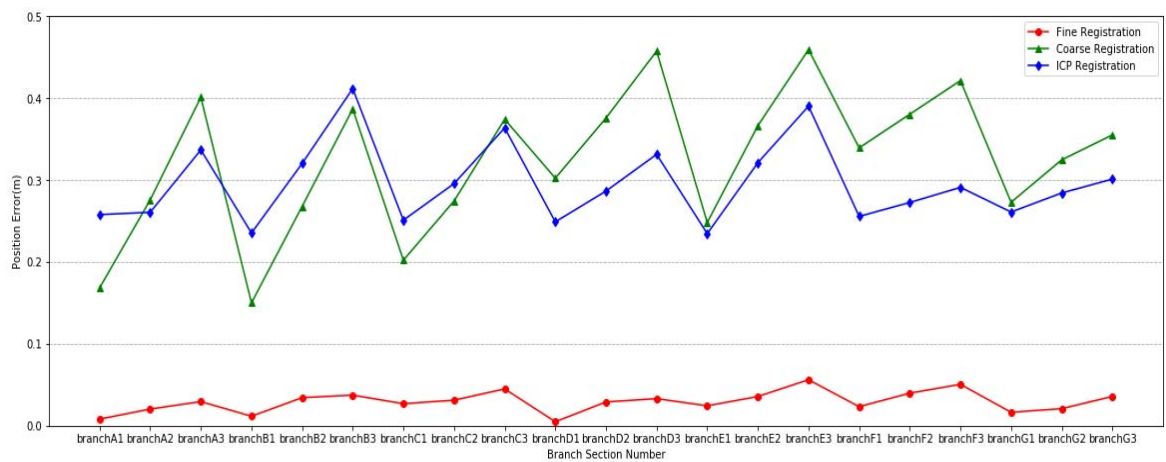

(b)

Fig. 14. Registration accuracy for simulated tree B. (a) Registration accuracy for scan 2 to scan 1. (b) Registration accuracy of scan 3 to scan 1.

These results demonstrate that fine registration errors were much smaller than those of coarse registration and that smaller fluctuations occurred among registration errors for branches. The ICP algorithm largely depends on the initial position of the point clouds, especially for adjacent scans with low-overlap areas. In most situations in our experiments, the ICP method exhibited no clear improvement compared with coarse registration, and registration errors increased after ICP fine registration. However, when coarse registration accuracy was relatively high, ICP fine registration enhanced registration accuracy, as shown by the registration results of scan 2 to scan 1 for simulated tree B (Fig. 14a).

3.4 Experiments on real-tree point cloud

Given that the structure of trees in nature scenes are more complex than that of simulated trees, the verification of our method on real-world tree point clouds is important. In the study, the a real-world tree point cloud is acquire by RIEGL VZ-400 TLS. RIEGL VZ-400 works in two modes: the long range mode and the high speed mode. In the long range mode, the maximum measuring distance is 600m, and the maximum measurement rate is 42000 measures/s. In the high speed mode the maximum measuring distance is 350m, and the maximum measurement rate is 122000 measures/s. The tree point cloud data is composed of 3 scans scanned from three different positions.(Fig.15) The result of coarse registration and fine registration are shown in Fig.16. To make the observation of the registration result more clear, we filter out the small branches and noisy leaf points and only showed the main structure of trees in Fig.16. To evaluate the registration accuracy, corresponding branches between adjacent scans were extracted based on our evaluation model. Branch section numbers are shown in Fig.17. The accuracies of the coarse and fine registration for the tree is shown in Figs.18. Table 3 shows the calculated mean registration errors.

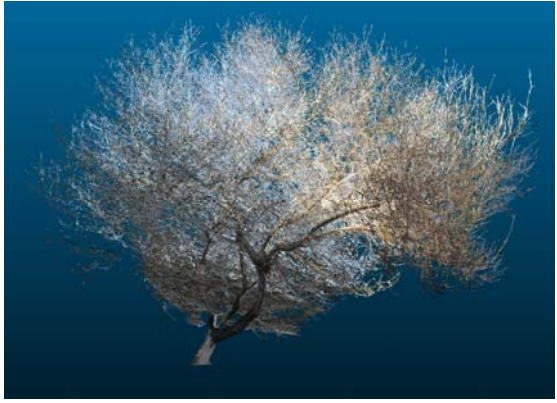
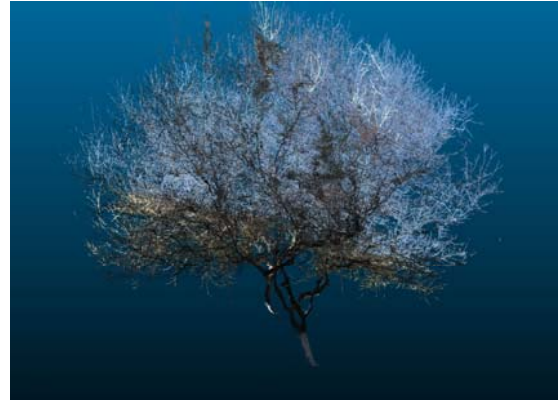

(a)  (b)

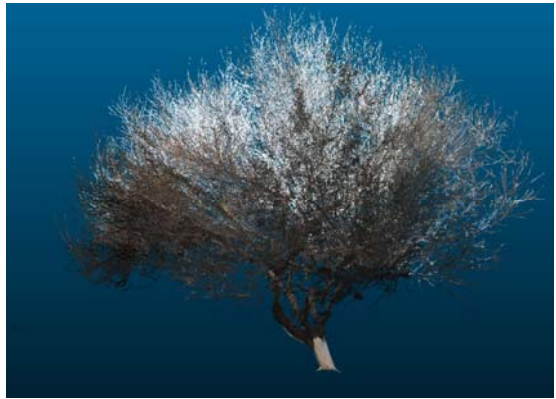

(c)

Fig. 15. Point cloud data of a real-world tree .(a) scan1. (b) scan2. (c) scan3

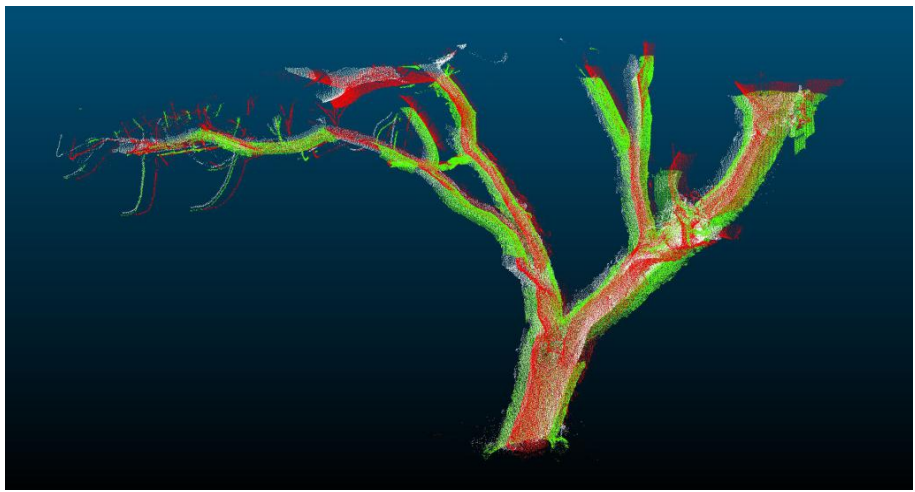

(a)

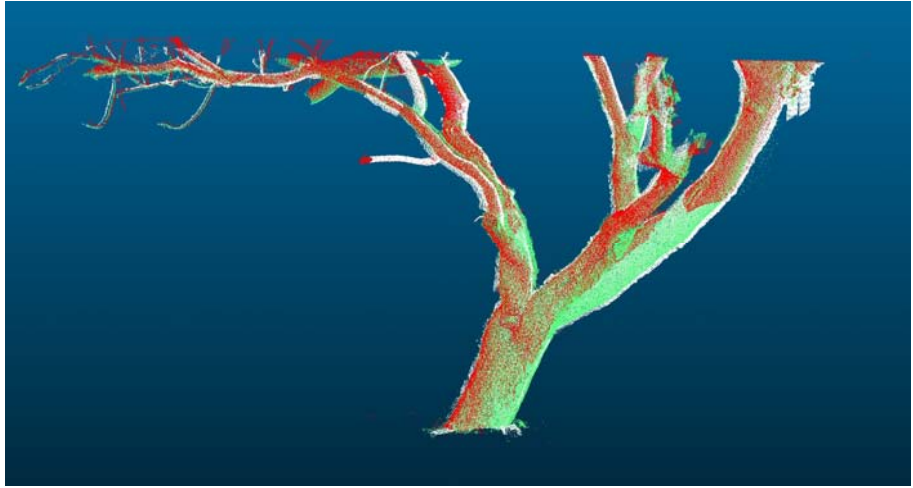

(b)

Fig. 16. Registration results. (a) Coarse registration results among 3 scans. (b) Fine registration results among 3 scans. The registration order is scan 2 to scan 1, scan 3 to scan 1. Green, red, and white points indicate scans 1, 2, and 3, respectively.

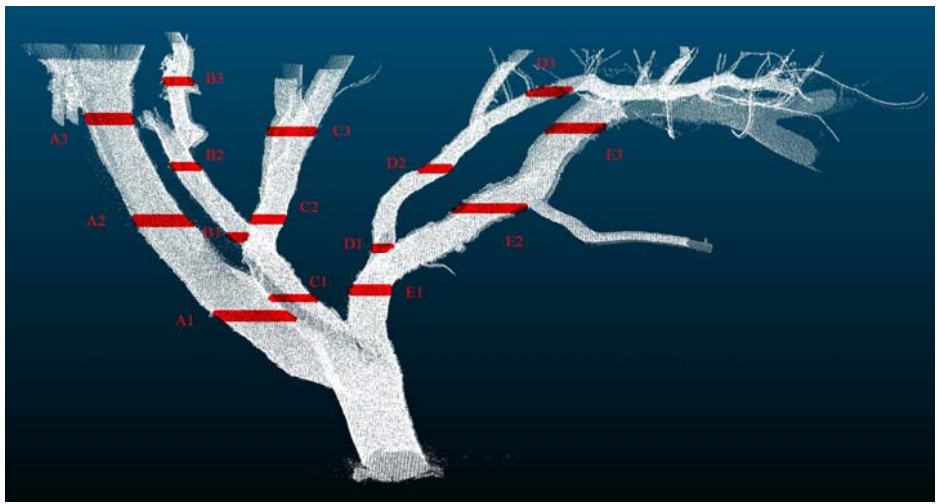

Fig. 17. Branch section numbers in the tree point cloud. Red, sliced branch parts.

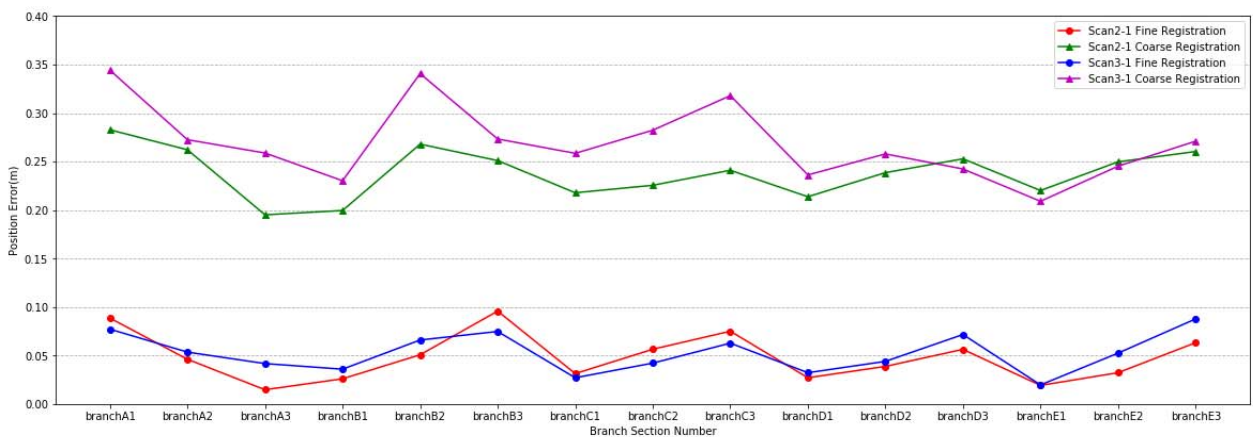

Fig. 18. Registration accuracy for the tree point cloud..

Table 3

The errors of registration of the tree point cloud.

| Registration errors(m) | Coarse registration | Fine registration |
|---|---|---|
| Scan 2-1 | 0.238 | 0.047 |
| Scan 3-1 | 0.269 | 0.052 |

Note : 2-1 means the registration order "Scan 2 to Scan 1", 3-1 means the registration order "Scan 2 to Scan 1".

The results show that the coarse registration error of real-world tree data is close to the coarse registration error of simulated trees. However, the fine registration error increases in experiments on real-world tree data.

**4 Discussion**

4.1 Verification of feature-point matching

In the proposed algorithm, coarse registration provided the initial position of the point cloud, which played an important role in the success of fine registration. In coarse registration, the matching accuracy of similar images determines tie-point quality and directly affects coarse-registration accuracy. However, due to the lack of texture information in binary images, wrong matching points cannot be avoided in feature-point matching (Fig. 9).

With fewer pairs of matching points, the impact of wrong matching points on the estimation of the transformation matrix between adjacent scans is greater. To enhance matching accuracy, verification of point pairs is performed to eliminate bad matches. Each pair of points ($P_1$, $P_2$) corresponds to a pair of intervals of angles α and β, which corresponds to the 3D space $\sigma_1$, $\sigma_2$. In a correct matching-point pair, the relative position of $\sigma_1$ in the tree point-cloud $PC_1$ is roughly equivalent to the relative position of $\sigma_2$ in the tree point-cloud $PC_2$. Given that the tree is composed of different parts in the vertical direction and can be roughly divided into the crown, limb, and trunk from top to bottom, the correct matching-point pair should be in the same part of the tree. For example, a point in the crown should be matched to a point in the crown. Thus, the correctness of matching-point pairs can be verified by comparing their corresponding intervals of angle β, which indicate their position in the vertical direction.

Suppose that each pair of matching points corresponds to the intervals [$\beta_{1k}$, $\beta_{1k}$ + 2$\phi$] and [$\beta_{2k}$, $\beta_{2k}$ + 2$\phi$] ($k$ = 1, 2, …, 15), where 15 is the number of pairs of matching points to be checked. We can then calculate the distance $d_k = |\beta_{1k} - \beta_{2k}|$ and obtain the mean value $\overline{d_k}$ and standard deviation $\sigma_d$ of $d_k$. A point pair whose $d_k$ value is not within $\left(\overline{dk} - \sigma_d, \overline{dk} + \sigma_d\right)$ will be excluded as a bad pair from potential matching points.

4.2 Improvement of point separation

The separation of fine-registration sliced points can be influenced by tree structure. In this study, points were sliced at quartiles of tree height. Fig. 19. shows an example in which parts of the trunk and branches are close at these specific heights. Due to the small distances between branches at height Q2, the sliced points of different branches intersected. Our method these considered intersected sliced points as a single connected part. Therefore, the corresponding sliced points could not be correctly separated (Fig. 19b).

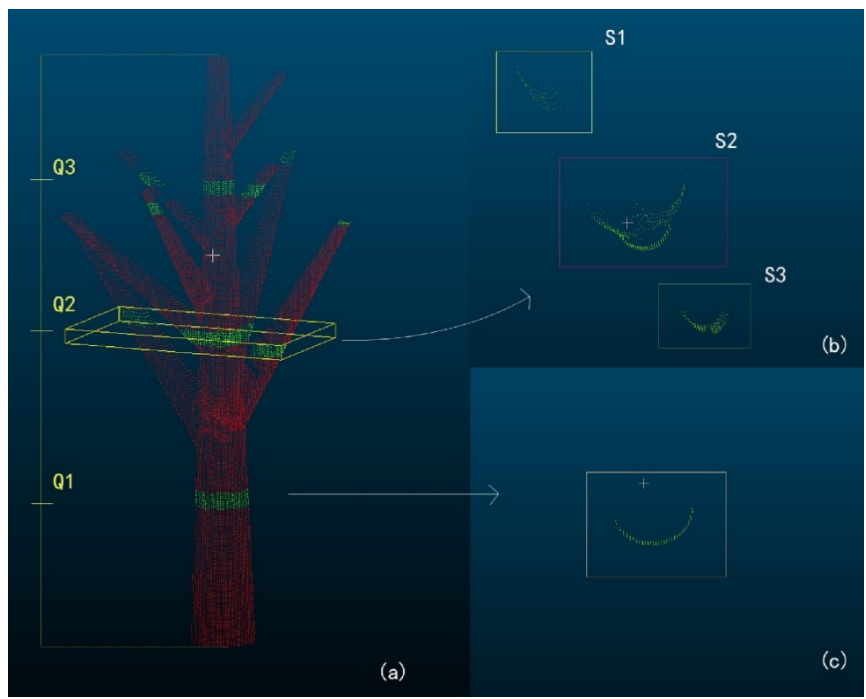

Fig. 19. Point-cloud slicing and separation results. (a) Point-slicing results. Q1, Q2, and Q3 are quartiles of tree height. (b) Separation of points sliced at Q2. (c) Separation of points sliced at Q1.

One method to improve our approach is to verify the correctness of separated parts. Points in rectangle S2 in Fig. 19b contained three arcs, which roughly compose an arc with the largest radius among all arcs. In most single trees, the thickness of trunks and branches tends to decrease as height increases, such that lower parts of the trunk or branches are often thick and higher parts are thinner. Therefore, the radius of a lower part of a trunk or branch should be smaller than that of a higher part. According to this rule, we can verify the correctness of a separated part by checking its corresponding radius after fitting. For example, by applying a fitting method, we obtained radius $r_1$ of the separated part shown in Fig. 19c and radius $r_2$ of separated part S2 in Fig. 19b. As $r_2$ was larger than $r_1$, S2 was considered an incorrect result and was excluded from further registration procedures.

## 5 Conclusion

The main objective of this study was to achieve registration of a single tree point-cloud from multiple scans without the aid of artificial reflectors. We proposed an automatic registration method that uses a coarse-to-fine strategy to register multiple scans of a single tree. Unlike methods that use reflectors as references to adjust point-cloud positions in coarse registration, our method depends on a change in dimension and extracts natural features of the tree to estimate transformation. Coarse registration uses a projection of the 3D point cloud to generate 2D images and apply a feature-point-matching algorithm to extract matching points. In fine registration, slicing, separation, and point fitting are applied to extract corresponding center points of the trunk and branches for use as tie points to calculate accurate rigid-body transformation parameters. Experiments using the proposed method were first conducted based on the data of two simulated trees and the results were compared to those using the ICP registration method. The registration error of our method was less than 0.03 m. The experiments on the real-world tree point cloud data further verify the effeteness of the method on real tree data.The registration error of the method is around 0.05 m.

There were several limitations to this study. To ensure that similar image pairs could be detected from corresponding images in adjacent scans, point clouds were rotated continuously in 3D space prior to projection. When the point cloud is large, this rotation will be time-consuming. The number and degree of rotations must be further optimized to reduce redundancy. In addition, feature-point-matching results were not stable, especially for trees with complex geometric structures. When there are many asymmetric structures in a tree, it will be challenging to extract sufficient correct matching-point pairs. Finally, fine registration relies on the fitting of branch centers. In a natural forest, some tree branches are slim and branch density is very high. The separation of branch parts can be difficult, and cylinder fitting of the slice points will not be accurate.

## Acknowledgments

This research is funded by the Fundamental Research Funds for the Central Universities (No.2015ZCQ-LY-02), National Students' innovation and entrepreneurship training program (No. 201710022076) and the State Scholarship Fund from China Scholarship Council (CSC No. 201806515050) .